\documentclass[sigconf]{acmart}
\AtBeginDocument{%
  }

\copyrightyear{2026}
\acmYear{2026}
\setcopyright{cc}
\setcctype{by}
\acmConference[WWW '26] {Proceedings of the ACM Web Conference 2026}{April 13--17, 2026}{Dubai, United Arab Emirates.}
\acmBooktitle{Proceedings of the ACM Web Conference 2026 (WWW '26), April 13--17, 2026, Dubai, United Arab Emirates}
\acmISBN{979-8-4007-2307-0/2026/04}
\acmDOI{10.1145/3774904.3792161
}

\usepackage{newfloat}
\usepackage{listings}

\usepackage{booktabs} 
\usepackage{graphicx} 
\usepackage{diagbox} 
\usepackage{multirow} 
\usepackage{amsmath}

\usepackage[utf8]{inputenc} 
\usepackage[T1]{fontenc}    
\usepackage{hyperref}       
\usepackage{url}            
\usepackage{amsfonts}       
\usepackage{nicefrac}       
\usepackage{microtype}      
\usepackage{xcolor}         
\usepackage{caption}
\usepackage{tabularx}

\usepackage{rotating}
\usepackage{wrapfig}
\usepackage{makecell}
\usepackage{float}
\usepackage{array}
\newcolumntype{L}{>{\raggedright\arraybackslash}X}
\newcolumntype{R}{>{\raggedright\arraybackslash}X}

\usepackage{fvextra}
\usepackage{tcolorbox}
\usepackage{tikz}
\usepackage{dashrule}  
\usepackage{lipsum}

\usepackage{enumitem}

\usepackage[ruled,vlined,linesnumbered]{algorithm2e}
\SetKwInOut{KwIn}{Input}
\SetKwInOut{KwOut}{Output}
\SetKw{KwContinue}{continue}
\usepackage{placeins}

\tcbuselibrary{breakable}
\setlist[itemize]{leftmargin=*}
\setlist[enumerate]{leftmargin=*}

\begin{document}

\title{Are LLMs Reliable Translators of Logical Reasoning Across Lexically Diversified Contexts?}
\title{Are LLMs Stable Formal Logic Translators in Logical Reasoning Across Linguistically Diversified Texts?}

\author{Qingchuan Li}
\orcid{0009-0009-9747-0888}
\affiliation{%
  \institution{State Key Laboratory of Cognitive Intelligence, University of Science and Technology of China}
  \city{Hefei}
  \state{}
  \country{China}
}
\email{chouli@mail.ustc.edu.cn}

\author{Jiatong Li}
\orcid{0009-0000-8877-6927}
\affiliation{%
  \institution{State Key Laboratory of Cognitive Intelligence, University of Science and Technology of China}
  \city{Hefei}
  \state{}
  \country{China}
}
\email{cslijt@mail.ustc.edu.cn}

\author{Zirui Liu}
\orcid{0009-0002-7263-9607}
\affiliation{%
  \institution{State Key Laboratory of Cognitive Intelligence, University of Science and Technology of China}
  \city{Hefei}
  \state{}
  \country{China}
}
\email{liuzirui@mail.ustc.edu.cn}

\author{Mingyue Cheng}
\orcid{0000-0001-9873-7681}
\affiliation{%
  \institution{State Key Laboratory of Cognitive Intelligence, University of Science and Technology of China}
  \city{Hefei}
  \state{}
  \country{China}
}
\email{mycheng@ustc.edu.cn}

\author{Yitong Zhou}
\orcid{0009-0007-6579-1092}
\affiliation{%
  \institution{State Key Laboratory of Cognitive Intelligence, University of Science and Technology of China}
  \city{Hefei}
  \state{}
  \country{China}
}
\email{yitong.zhou@mail.ustc.edu.cn}

\author{Yuting Zeng}
\orcid{0009-0006-1719-5424}
\affiliation{%
  \institution{School of Computer Science and Technology, University of Science and Technology of China}
  \city{Hefei}
  \state{}
  \country{China}
}
\email{yuting_zeng@mail.ustc.edu.cn}

\author{Qi Liu}
\authornote{Qi Liu is the corresponding author.}
\orcid{0000-0001-6956-5550}
\affiliation{%
  \institution{State Key Laboratory of Cognitive Intelligence, University of Science and Technology of China}
  \city{Hefei}
  \state{}
  \country{China}
}
\email{qiliuql@ustc.edu.cn}

\author{Tongxuan Liu}
\orcid{0009-0007-2634-2788}
\affiliation{%
  \institution{School of Information Science and Technology, University of Science and Technology of China}
  \city{Hefei}
  \state{}
  \country{China}
}
\affiliation{%
  \institution{JD.com}
  \city{Beijing}
  \state{}
  \country{China}
}
\email{tongxuan.ltx@mail.ustc.edu.cn}

\renewcommand{\shortauthors}{Qingchuan Li et al.}

\begin{abstract}

Logical reasoning with large language models (LLMs) has received growing attention. One mainstream approach translates natural language into formal logic and applies symbolic solvers for deduction. However, while effective in many tasks, LLM-based translators often fail to produce consistent symbolic representations when the same concept appears in different linguistic forms. Such inconsistencies break logical coherence and cause solver errors. Moreover, most existing benchmarks lack this type of linguistic variation, which is common in real-world text, leaving the problem underexplored.
To address this gap, we present SoLT, a benchmark that systematically rewrites reasoning datasets into diverse yet logically equivalent forms at multiple levels. Beyond evaluation, SoLT also offers a general method to enrich datasets with linguistic diversity while preserving meaning and logic.
To further enhance the stability of LLM-based reasoning, we propose MenTaL, which explicitly guides models to construct a concept–symbol mapping table during translation. By linking equivalent expressions to shared symbols, MenTaL maintains consistency and mitigates symbol drift. Experiments on SoLT demonstrate that LLMs suffer from inconsistent symbol mapping under linguistic variation, leading to drops in reasoning accuracy. Meanwhile, applying MenTaL yields clear and stable performance gains across diverse inputs.
Overall, our findings reveal that overlooking linguistic diversity obscures key weaknesses in LLM-based translators, and our work takes a step toward more reliable logical reasoning in real-world scenarios.
Our code is available at 
\url{https://github.com/wufeiwuwoshihua/LinguDiver}.

\end{abstract}



\maketitle

\section{Introduction}

Logical reasoning is fundamental to both human cognition and artificial intelligence~\cite{ippoliti2020introduction,wang2025can,cheng2025can,pan2026paperscout,liu2024computerized}. 
Recently, neuro-symbolic approaches have become a mainstream paradigm for handling long reasoning chains~\cite{pan2023logic,olausson2023linc,luo2025time}. 
These approaches combine large language models with symbolic solvers: LLMs translate natural language into formal logic, and solvers perform deduction.
Since symbolic solvers are logically complete~\cite{prover9, z3, pyke, de2008z3}, system reliability largely depends on the translation step~\cite{ryu2024divide,li2025hypothesis}, making faithful and consistent natural-language–to–logic translation essential.

\begin{figure}[t]
  \centering
  \includegraphics[width=0.98\columnwidth]{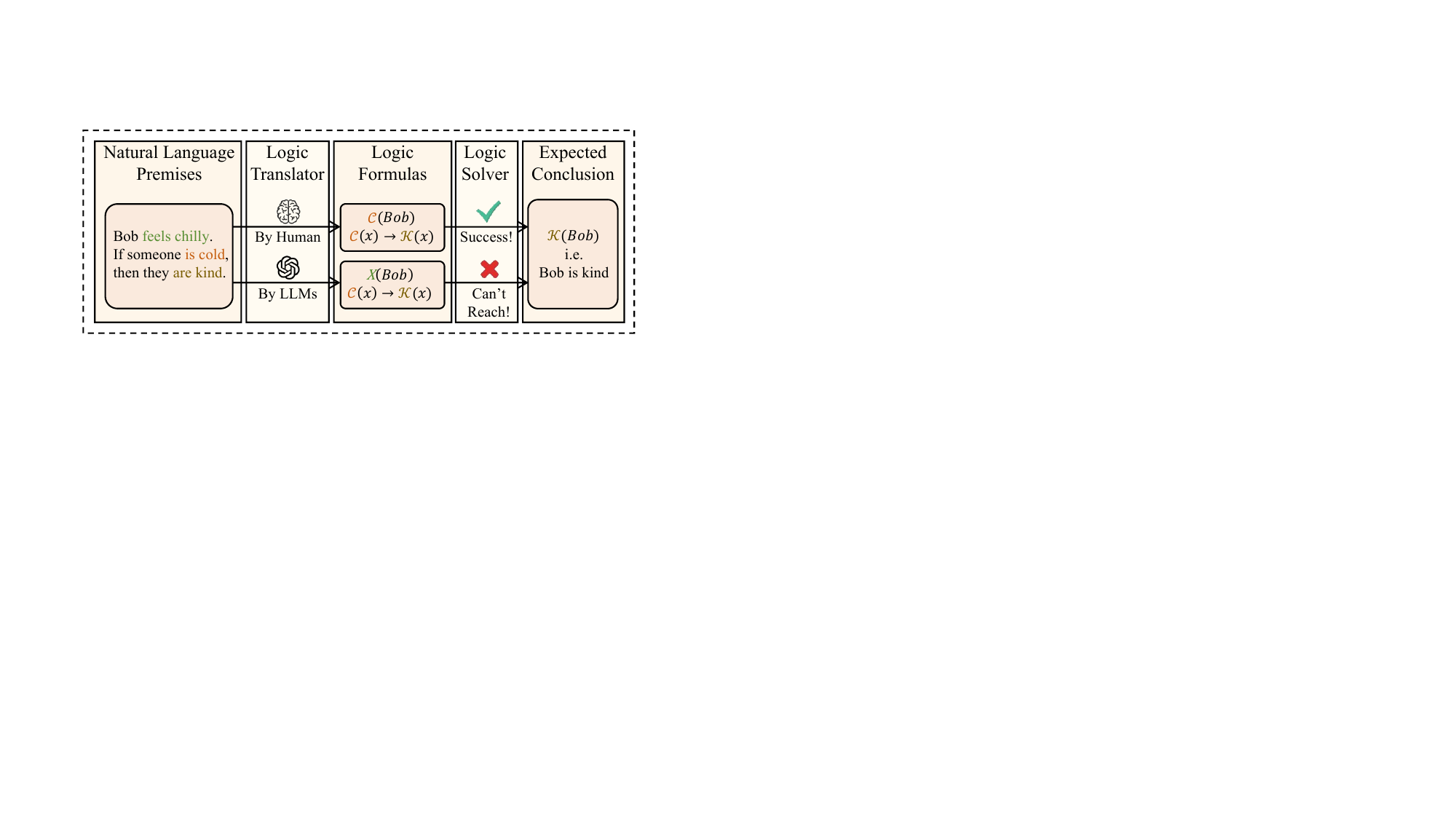}
  \caption{An illustrative case of LLM translators failing under linguistic diversification. LLMs map equivalent expressions to different symbols, breaking reasoning chains and preventing the solver from reaching the correct conclusion.}
  \label{fig:intro}
\end{figure}

Since logical reasoning systems rely on accurate translation, the translator must preserve semantics and ensure consistent symbol mapping, where semantically equivalent expressions share the same logical symbol~\cite{olausson2023linc}.
Symbolic solvers operate purely on symbols, so inconsistent mappings disrupt the reasoning process.
This issue, known as \emph{symbol drift}, breaks the reasoning chain and prevents correct conclusions~\cite{prover9} (see Figure~\ref{fig:intro}).
This risk of symbol drift is heightened by the fact that real-world language often expresses the same concept through synonyms, paraphrases, or syntactic variations~\cite{williams2014lessons}.
Therefore, consistent symbol mapping across linguistic variation is essential for reliable logical translation in practice.


Despite its importance, existing reasoning benchmarks fail to effectively assess models’ ability to maintain consistent symbol mapping~\cite{saparov2022language,tafjord2020proofwriter,han2022folio}.
These benchmarks primarily emphasize logical structure and reasoning depth, while their language remains highly uniform, often using identical expressions for repeated concepts.
As a result, they cannot systematically evaluate translation stability under realistic linguistic variation, and are therefore likely to overestimate the robustness of LLM-based logical translation.
To verify the practical impact of this evaluation gap, we conduct exploratory experiments that introduce mild, semantics-preserving linguistic variations.
The results reveal a clear drop in translation accuracy, indicating that even small language changes can disrupt symbol consistency.
These findings highlight the need for a benchmark that explicitly includes linguistic diversity to more reliably evaluate and improve LLM-based logical translation.

To address this issue, we introduce \textbf{SoLT}, a benchmark for evaluating the stability of logical translation under linguistic diversity.
SoLT is built from existing reasoning datasets using a \textbf{logic-invariant linguistic diversification} method that generates linguistically diverse but logically equivalent problem variants.
Inspired by how humans naturally vary expressions in communication~\cite{williams2014lessons,bucholtz2015elements,roget2023thesaurus}, the method introduces controlled variation across words, phrases, and sentence forms, while preserving semantic and logical consistency.  
Empirical validation shows that this diversification substantially increases linguistic variation without altering semantics or logic.
Consequently, SoLT enables reliable evaluation of translation stability and provides a general way to enrich reasoning datasets with realistic linguistic variation.

While SoLT exposes the instability of LLM-based translation under linguistic diversity, achieving stable translation remains a non-trivial challenge.
To tackle this issue, we propose \textbf{MenTaL}, a mental representation table–guided framework for formal logic translation that mitigates symbol drift at the concept level.
Inspired by theories of mental representation~\cite{carey2009our,fodor1975language,pinker2007stuff,frege1892sense}, MenTaL explicitly organizes semantically equivalent expressions within a mental representation table prior to logical symbol assignment, thereby promoting consistent symbol mapping under linguistic variation.
The framework accommodates both prompt-based adaptation for large closed-source models and supervised fine-tuning for smaller open-source models, enabling broad applicability across diverse experimental settings.

Through experiments on SoLT, we confirm that current LLMs indeed suffer significant performance drops under diversified inputs and exhibit pervasive symbol drift, revealing their difficulty in maintaining consistent symbol alignment.  
Further analyses show that SoLT can effectively transform existing reasoning benchmarks to better diagnose weaknesses in LLM-based translators.  
We also demonstrate that MenTaL can mitigate this problem, consistently improving translation accuracy across tasks and models, offering insight for future research on stable logical translation.



Our contributions are as follows:

\begin{itemize}
    \item We propose SoLT, a benchmark that evaluates the stability of logical translation under linguistic diversity, constructed via a general method that rewrites reasoning datasets into diverse yet logically equivalent forms.
    \item We introduce MenTaL, a lightweight and generally applicable framework that guides LLMs to build a concept–symbol mapping table during translation, reducing symbol drift and improving the stability of the translation process.
    \item Extensive experiments on SoLT reveal symbol drift in current LLM translators and demonstrate that MenTaL effectively improves their robustness and stability.
\end{itemize}
\section{Related Works}

\paragraph{Natural Language Datasets for Logical Reasoning.}
Existing datasets span a wide range of reasoning tasks \cite{qi2025large}, and many of them provide explicit reasoning chains \cite{saparov2022language}.
However, a large portion of these benchmarks \cite{tafjord2020proofwriter,qi2025large,saparov2022language,ghazal2013bigbench} are generated using rule-based templates, which tend to produce repetitive surface realizations of the same underlying logical concepts.
Other datasets \cite{han2022folio,yu2020reclor,liu2020logiqa,wang2022lsat} are manually curated or derived from standardized exams; while they emphasize reasoning difficulty and coverage of logical rules, they pay limited attention to linguistic variation.
As a result, recurring expression patterns dominate these benchmarks, preventing them from reflecting the diversity of language commonly observed in real-world reasoning scenarios.
Consequently, they are ill-suited for evaluating whether LLMs can consistently assign symbols to semantically equivalent expressions during logical translation.
To address this gap, our work introduces a linguistic diversification method built on top of existing datasets, serving as a complementary resource to current reasoning benchmarks.


\paragraph{Adversarial Text Attacks on LMs.}  
Adversarial text attacks assess model robustness by introducing small, semantically preserving perturbations that induce misleading behavior.
Prior work spans jailbreak attacks~\cite{wei2023jailbroken,souly2024strongreject,chao2024jailbreakbench,xue2025dual} and adversarial text classification~\cite{xu2023llm}, with methods operating at the character~\cite{gao2018black}, word~\cite{li2023adversarial,jin2020bert}, or sentence level~\cite{lin2021using,zhang2024careful}, often via synonym substitution~\cite{jin2020bert,rauba2024quantifying}.
More recent approaches leverage LLM-based frameworks such as PromptAttack~\cite{xu2023llm,liu2025elo}.
However, existing methods do not address symbol consistency in logical translation.
In contrast, Our work introduces a new paradigm, logic-invariant linguistic diversification, which extends adversarial methods to systematically test semantic stability in logical reasoning tasks.  


\paragraph{LLMs as Formal Logic Translators.}  
Recent studies view LLMs as translators that convert natural language reasoning problems into formal logic, solved either by symbolic solvers \cite{olausson2023linc, ye2023satlm, pan2023logic, kirtania2024logic} or by the models themselves \cite{xu2023symbol, li2024leveraging}.
This line of work positions LLMs as a bridge between natural language and formal logic, advancing automated reasoning.
However, most existing methods focus on local translation accuracy, such as parsing and translating logical relations within single sentences \cite{ryu2024divide, kirtania2024logic}, while overlooking global \textbf{symbol consistency}.
As a result, they fail to ensure that semantically equivalent concepts are mapped to the same logical symbols throughout the reasoning process.

\section{Experimental Explorations}

To examine whether linguistic diversity undermines the stability of LLM-based logical translation even when semantics and logic are preserved, we conduct an exploratory experiment. We manually introduce high-quality linguistic variations into existing reasoning benchmarks while keeping their semantics and reasoning structure unchanged. Each sample is modified with at most two perturbations from four common types: 
(1) \textbf{Third-person reference}, replacing explicit mentions with pronouns or indirect descriptions; 
(2) \textbf{Synonym substitution}, replacing words with semantically equivalent alternatives; 
(3) \textbf{Part-of-speech shift}, converting words across grammatical categories; and 
(4) \textbf{Syntactic transformation}, modifying phrase or clause structures (e.g., active–passive alternation). This setting enables a focused evaluation of translation stability under natural yet minimal linguistic variation.

For evaluation, we follow the task–solver pairs and prompt settings of Logic-LM~\cite{pan2023logic}. GPT-4 is used as the translator, and symbolic solvers assess the correctness of the resulting logical forms. We report the accuracy drop before and after linguistic diversification. Experimental details are provided in Appendix~\ref{app:exp_set}.

The results in Figure~\ref{fig:heat} show consistent performance degradation across all variation types.
\textbf{Third-person reference} causes the most severe decline (about $0.27$ on average, up to $0.33$ on ProntoQA), indicating difficulty in linking indirect mentions to consistent logical predicates.
Synonym substitution also leads to substantial errors (mean drop $\approx 0.22$), while Part-of-speech shift and Syntactic transformation have milder but still non-negligible effects (around $0.13$ and $0.11$).
Performance drops are more evident on complex datasets such as FOLIO~\cite{han2022folio} and ProverQA~\cite{qi2025large}.

Inspection of translated outputs shows that these failures mainly arise from inconsistent symbol assignment: semantically equivalent expressions are mapped to different logical symbols (e.g., \textit{student} vs.\ \textit{the pupil}), breaking logical equivalence and causing solver errors.
Notably, even mild linguistic variation leads to clear accuracy degradation, suggesting that benchmarks without linguistic diversity substantially overestimate the stability of LLM-based logical translators. This motivates the need for a benchmark that explicitly evaluates translation under realistic linguistic variation.

\section{Methodology}

\begin{figure}[t]
    \centering
    \includegraphics[width=.9\linewidth]{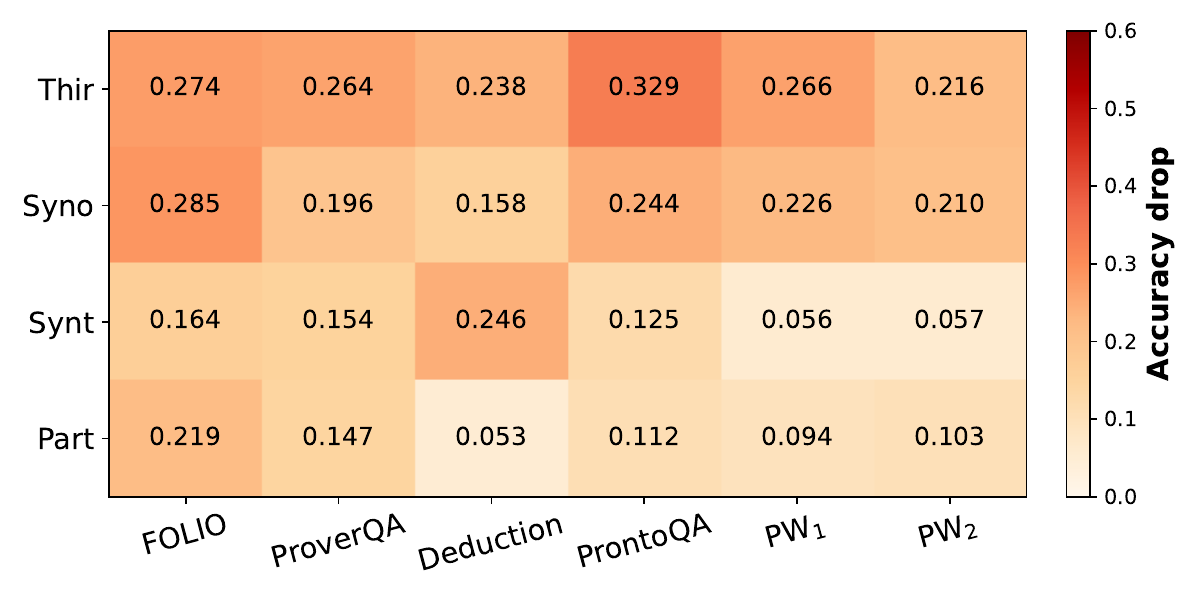} 
    \caption{Accuracy drop of GPT-4 across six reasoning tasks under four diversification types: third-person reference (Thir), synonym substitution (Syno), part-of-speech shift (Part), and syntactic transformation (Synt).}
    \label{fig:heat}
\end{figure}

\begin{figure*}[t]
  \centering
  \includegraphics[width=0.98\textwidth]{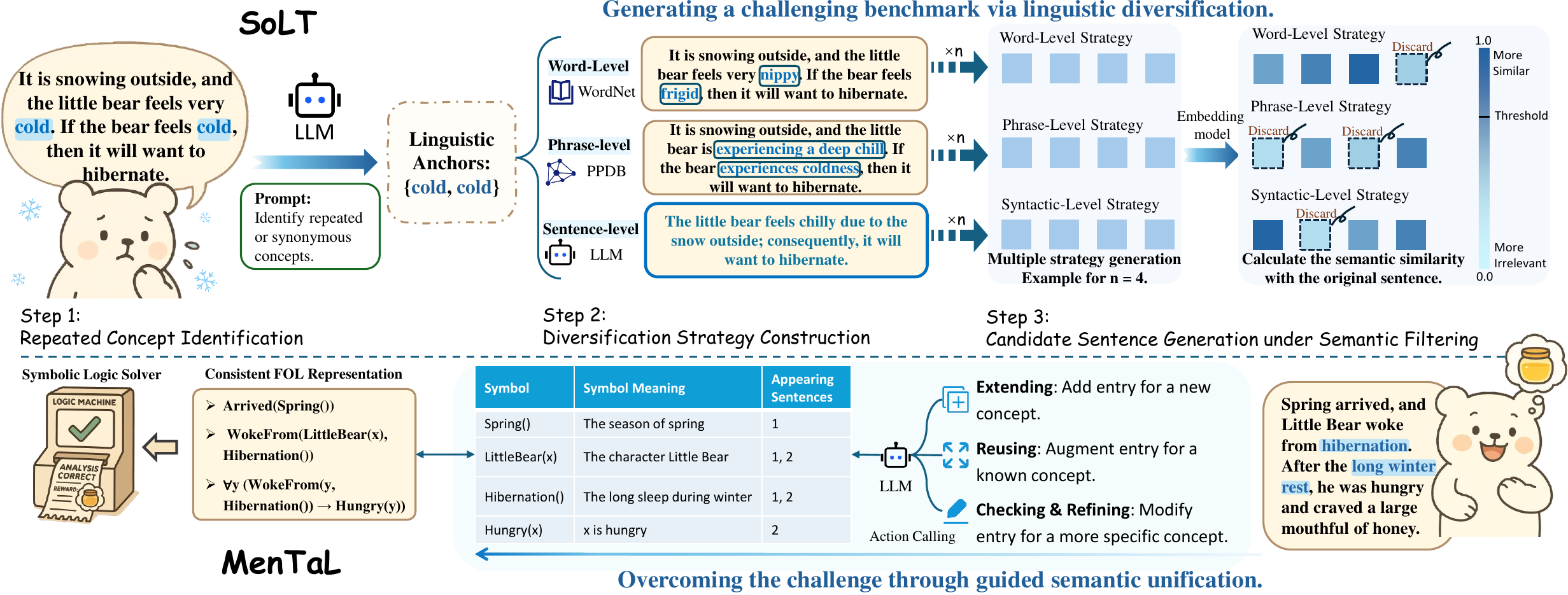}
  \caption{SoLT performs multi-level linguistic diversification while preserving semantics and logic to evaluate translation stability. MenTaL enforces unified symbol mapping through a Mental Representation Table.}
  \label{fig:main}
\end{figure*}

\subsection{Logic-invariant Linguistic Diversification}

This section introduces \textbf{Logic-invariant Linguistic Diversification}, a method that systematically increases surface-level linguistic variation in reasoning questions while preserving their original semantics and logical structure.
Formally, given an input problem
\[
q = \{s_1, s_2, \dots, s_n\},
\]
where \(s_i\) denotes a sentence, we construct a diversified counterpart
\[
q' = \{s_1', s_2', \dots, s_n'\},
\]
in which \(s_i'\) is one of rewritten form of \(s_i\) that maintains semantic and logical equivalence. 

To implement this idea, we design a structured diversification pipeline that balances linguistic diversity with logical fidelity.
As illustrated in the upper part of Figure~\ref{fig:main}, the pipeline consists of three stages: identifying repeated concepts, constructing diversified alternatives, and generating semantically filtered sentences.
We describe these stages in detail below.



\textbf{Step 1: Repeated Concept Identification.}
In logical reasoning, recurring concepts connect multiple sentences into a coherent inference chain, yet existing benchmarks often reuse identical surface expressions to refer to the same concept, masking challenges in handling linguistic variation.
To identify candidates for diversification, we detect surface expressions that appear multiple times within an input problem.
Formally, let \(e\) denote a surface expression and \(\mathrm{freq}_q(e)\) its frequency in \(q\).
We define the set of repeated expressions as
\[
C_{\text{rep}}(q) = \{\, e \mid \mathrm{freq}_q(e) \ge 2 \,\},
\]
which serve as anchors for subsequent diversification.




\textbf{Step 2: Diversification Strategy Construction.}
We construct fine-grained diversification strategies for repeated concepts using a three-level \emph{parallel} scheme that reflects natural linguistic variation at the word, phrase, and sentence levels.
For each repeated expression \(e \in C_{\text{rep}}(q)\), we generate candidate variants via:
\begin{itemize}
    \item \textbf{Word level:} synonym substitution;
    \item \textbf{Phrase level:} paraphrasing short phrases;
    \item \textbf{Sentence level:} controlled syntactic rewrites (e.g., voice alternation or third-person reformulation).
\end{itemize}

WordNet~\cite{princeton2010aboutwordnet} and PPDB~\cite{ganitkevitch2013ppdb} are used for word- and phrase-level variants, while LLMs are applied only for sentence-level rewrites or as a fallback when lexical resources yield no valid candidates.
All candidates are subsequently filtered by contextual semantic similarity in Step~3 to prevent semantic drift.

\textbf{Step 3: Candidate Sentence Generation under Semantic Filtering.}
This stage applies the diversification strategies from Step~2 to generate diversified sentences while preserving the original meaning, and filters out candidates that introduce semantic drift. 
For each sentence \(s_i \in q\), which may contain one or more repeated expressions \(e \in C_{\text{rep}}(q)\), we replace these expressions with candidates generated in Step~2, yielding a set of candidate sentences \(\{s_i^{(k)}\}\).
Since direct replacement may alter sentence meaning, we assess semantic consistency using strong sentence embedding models~\cite{qwen3embedding,bge_embedding} that capture sentence-level semantics in context, and retain only candidates whose similarity to \(s_i\) exceeds a threshold \(\theta\).
Formally, the filtered candidate set for \(s_i\) is defined as
\[
\mathcal{S}_i = \{\, s_i^{(k)} \mid \mathrm{sim}(s_i, s_i^{(k)}) \ge \theta \,\},
\]
ensuring that retained candidates differ from the original sentence mainly in lexical or syntactic form rather than meaning. 

After semantic filtering, multiple candidates may remain for each sentence, and many of them can share identical surface expressions for the same repeated concept across different sentences. 
To maximize surface diversity at the problem level, we employ an LLM to assemble the final diversified problem by selecting one candidate per sentence from the filtered sets, while minimizing repeated surface expressions for identical concepts. 
The final diversified problem is constructed as
\[
q' \in \mathcal{S}_1 \times \mathcal{S}_2 \times \cdots \times \mathcal{S}_n.
\]
During this selection process, the LLM selects only from the filtered candidates and prefers combinations that use diverse surface forms for the same concept across sentences.

Using this logic-invariant diversification procedure, we build SoLT versions of datasets to evaluate the stability of logical translation under controlled linguistic variation.

\begin{table}[t]
\centering
\small
\caption{Semantic preservation of SoLT datasets measured by embedding similarity and logical invariance scores.}
\label{tab:logical-invariance-scores}
\resizebox{.9\linewidth}{!}{
\begin{tabularx}{0.92\linewidth}{@{}l*{5}{>{\centering\arraybackslash}X}@{}}
\toprule
& \multicolumn{2}{c}{\textbf{Similarity}} & \multicolumn{3}{c}{\textbf{Score}} \\
\cmidrule(lr){2-3}\cmidrule(l){4-6}
\textbf{Dataset} & \textbf{Qwen3} & \textbf{BGE} & \textbf{Human} & \textbf{GPT-4} & \textbf{DS-R1} \\
\midrule
FOLIO        & 0.8928    & 0.9339    & 4.5 & 4.7 & 4.6 \\
ProverQA     & 0.9013    & 0.9436    & 4.6 & 4.9 & 4.8 \\
ProntoQA     & 0.9259    & 0.9610    & 4.9 & 4.2 & 4.8 \\
Deduction    & 0.8771    & 0.9231    & 4.3 & 4.5 & 4.4 \\
ProofWriter  & 0.9309    & 0.9484    & 4.9 & 4.8 & 4.9 \\
\bottomrule
\end{tabularx}}
\end{table}

\subsection{Validation of Logical Invariance}

To assess whether linguistic diversification alters the meaning of the data, we first evaluate \emph{semantic preservation} using embedding models, which are widely used to measure semantic similarity between texts.
Specifically, given pairs of original and diversified questions,
\[
D_{\text{dual}} = \{(q_i, q_i') \mid i = 1, 2, \ldots, |D|\},
\]
we compute similarity scores as a coarse-grained indicator of semantic drift.
We use two sentence embedding models, \textbf{Qwen3-embedding-8B} and \textbf{BGE-large-en-v1.5}.

As shown in Table~\ref{tab:logical-invariance-scores}, semantic similarity remains high across all five datasets after diversification with SoLT.
With Qwen3-embedding-8B, most scores exceed $0.90$ and none fall below $0.85$; with BGE-large-en-v1.5, all scores are above $0.90$.
These results suggest that SoLT introduces limited surface variation and largely preserves sentence-level semantics.

Building on this, we conduct a finer evaluation of logical invariance.
Following prior work on logical equivalence, we define a set of logical invariance criteria that consider both meaning and reasoning, as shown in Table~\ref{tab:logic-invariance}.

Following these criteria, we invited professional annotators and LLMs to evaluate the diversified data.  
A 1–5 SoLT was used to rate the quality of diversification:
\textbf{5.0} — fully satisfies all criteria and is entirely acceptable;
\textbf{4.0} — contains minor flaws but remains acceptable overall;
\textbf{3.0} — partially acceptable, with flaws that affect quality;
\textbf{2.0} — major flaws make the question unacceptable;
\textbf{1.0} — critical flaws make the question unusable.

The final score is computed as the average of human and model evaluations.
Inter-annotator agreement is high, with \textbf{Cohen's $\kappa = 0.83$}, indicating reliable human judgments.
As reported in Table~\ref{tab:logical-invariance-scores}, all diversified datasets score above $4.0$, and most exceed $4.5$, indicating strong preservation of both semantic content and logical structure.

Finally, to ensure that diversification does not affect human solvability, annotators also answer both original and diversified questions.
Detailed human evaluation results and annotation protocols are reported in Appendix~\ref{app:human}.

\subsection{MenTaL: A Mental Representation-Guided Translation Framework}

\begin{table}[t]
\centering
\caption{Standards for Logical Invariance Scoring.}
\label{tab:logic-invariance}
\resizebox{\linewidth}{!}{
\begin{tabular}{@{}p{1.5cm}p{7.5cm}@{}}
\toprule
\textbf{Name} & \textbf{Description} \\
\midrule
\makecell[tl]{Semantic\\ Invariance} &
The diversified question must preserve the same semantic information as the original. \\
\addlinespace
\makecell[tl]{Reasoning\\ Invariance} &
The reasoning process required to reach the answer must remain identical to that of the original, without introducing new premises. \\
\addlinespace
\makecell[tl]{Answer\\ Invariance} &
The answer obtained from the diversified question must be semantically equivalent to that of the original. \\
\addlinespace
\makecell[tl]{Statement\\ Clarity} &
The diversified question should be clearly expressed, without introducing confusion or ambiguity. \\
\bottomrule
\end{tabular}}
\end{table}

\begin{figure*}[ht]
    \centering
    \includegraphics[width=0.95\textwidth]{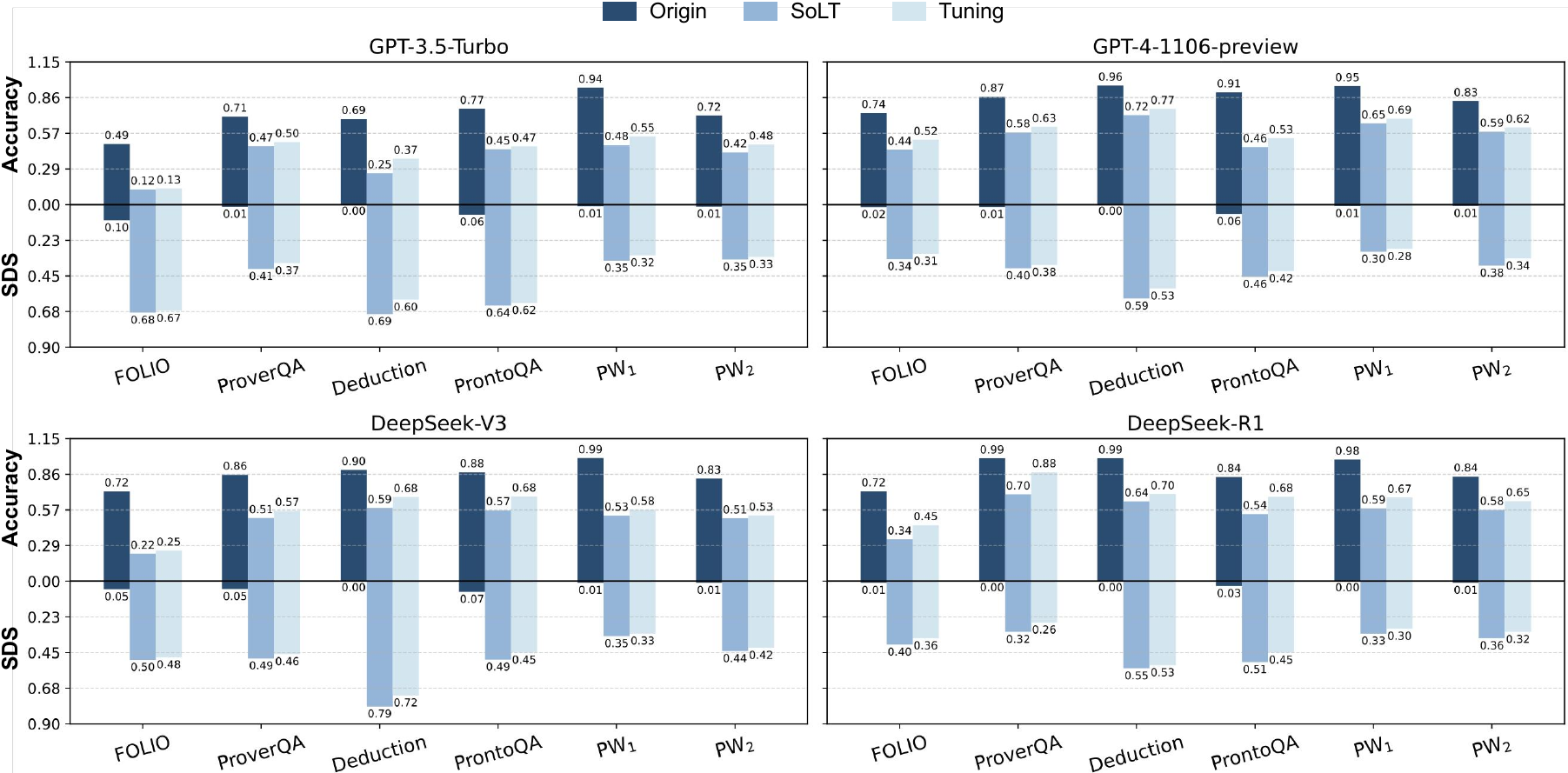} 
    \caption{Accuracy$\uparrow$ and SDS$\downarrow$ on six reasoning tasks across four LLMs under three evaluation settings. PW$_1$ and PW$_2$ denote the ProofWriter dataset paired with the Prover9 and PyKE solvers, respectively. Origin refers to the original dataset with direct prompting, SoLT refers to the SoLT-diversified dataset with direct prompting, and Tuning refers to the SoLT-diversified dataset with prompt-tuning. Accuracy is shown in the upper bars, while SDS is shown in the lower bars.}
    \label{fig:multi-model-accuracy}
\end{figure*}

As a further step, we introduce a lightweight approach, \textbf{MenTaL}, designed to mitigate symbol drift and enhance the stability of LLM-based logical reasoning systems.
Drawing inspiration from the theory of \emph{mental representation} in cognitive science~\cite{carey2009our,fodor1975language,frege1892sense}, MenTaL incorporates an explicit \emph{Mental Representation Table} (MRT) that the model maintains and updates throughout the translation process, as illustrated in the lower part of Figure~\ref{fig:main}.
The MRT functions as an external memory that records concept-level correspondences between expressions, ensuring consistent symbol assignment while preserving interpretability.

Formally, given a reasoning problem \(q\), MenTaL maintains a state \((\tilde{q}, T)\), where \(\tilde{q}\) is the evolving logical form and \(T\) is a table mapping semantically equivalent expressions to a unified symbol:
\[
T : \{e^{(1)}, e^{(2)}, \dots, e^{(m)}\} \mapsto \sigma.
\]
Here, \(\{e^{(i)}\}\) denotes expressions referring to the same concept, and \(\sigma\) is their shared logical symbol.
To update \(T\), MenTaL defines two relations between expressions:
\emph{semantic equivalence} (\(e^{(i)} \equiv e^{(j)}\)), indicating the same concept, and
\emph{conflict} (\(e^{(i)} \bowtie e^{(j)}\)), indicating partial overlap or subsumption (e.g., \emph{popular show} vs.~\emph{show}).

Based on these relations, MenTaL applies three update operations when a new expression \(e^{(i)}\) appears:
\begin{itemize}
    \item \textbf{Extending:} add a new entry for a previously unseen concept and assign it a new logical symbol.  
    \item \textbf{Reusing:} augment an existing entry when \(e^{(i)} \equiv e^{(j)}\) for some \(e^{(j)} \in T\), ensuring symbol consistency.  
    \item \textbf{Checking \& Refining:} modify an entry when \(e^{(i)} \bowtie e^{(j)}\) for some \(e^{(j)} \in T\), preserving the atomic base symbol while refining the mapping, e.g., \(\text{PopularShow}(x) \Rightarrow \text{Popular}(x) \wedge \text{Show}(x)\).  
\end{itemize}
Whenever \(T\) is updated, earlier parts of \(\tilde{q}\) are revised to maintain global symbol consistency.

MenTaL can be implemented through prompt-based interaction, where an LLM incrementally updates \(T\) during translation, or through supervised fine-tuning that jointly learns to produce \((\tilde{q}, T)\).  
The method does not aim to eliminate symbol drift entirely, but to substantially reduce it by enforcing explicit concept tracking.  
By making the mapping between language and logic transparent and auditable, MenTaL improves the stability and interpretability of LLM-based logical translation under linguistic variation.

\section{Evaluation of LLM Translators on SoLT}
\subsection{Experiment Setup}
\label{section:setup}

\paragraph{Solvers and Datasets.}
We evaluate LLM-based logical translation using three symbolic solvers: Python-Constraint, PyKE, and Prover9.
Five datasets covering diverse reasoning styles and complexities are used: LogicDeduction~\cite{ghazal2013bigbench}, ProntoQA~\cite{saparov2022language}\footnote{To ensure diversifiability, words in ProntoQA not attested in English are replaced with equivalent concepts from ProofWriter.}, FOLIO~\cite{han2022folio}, ProofWriter~\cite{tafjord2020proofwriter}, and ProverQA~\cite{qi2025large}.
Detailed dataset statistics and solver--dataset mappings are provided in Appendix~\ref{app:exp_set}.

\paragraph{Large Language Models.}
Experiments are conducted using four LLMs: GPT-3.5-Turbo~\cite{openai_gpt35}, GPT-4~\cite{achiam2023gpt}, DeepSeek-V3~\cite{liu2024deepseek}, and DeepSeek-R1~\cite{guo2025deepseek}.  
All models are evaluated in a 5-shot in-context setting.

\paragraph{Prompting Strategies.}
We consider two representative prompting strategies commonly used in logical translation:
(1) \textbf{Direct}, where the LLM translates natural language into formal logic using structured prompts with few-shot examples, following prior work such as Logic-LM~\cite{pan2023logic};
(2) \textbf{Prompt-Tuning}, which augments the prompt with explicit instructions and revised few-shot examples to encourage consistent symbol mapping across synonymous expressions.
Additional baseline results are reported in Appendix~\ref{appendix:a}.

\paragraph{Semantic Filtering Setting.}
We use the \texttt{bge-large-en-v1.5}~\cite{bge_embedding} sentence embedding model to compute cosine similarity between original and candidate sentences, retaining only candidates with similarity above a threshold $\theta = 0.90$.

\paragraph{Evaluation Metrics.} 
We report two metrics: \emph{Accuracy} and the proposed \emph{Symbol Dispersion Score (SDS)}.  

\textbf{Accuracy} measures the proportion of translated formulas that yield correct answers when executed by the solver. It directly reflects translation quality and serves as the main evaluation metric.

\textbf{SDS} quantifies symbol drift resulting from inconsistent semantic mapping.  
Let \(V\) be the set of semantic concepts in the dataset and \(\mathcal{L}\) the set of all logical symbols.  
A translation defines a mapping \(f: V \to 2^{\mathcal{L}}\), where each concept \(v \in V\) is associated with the set \(f(v)\) of symbols used to represent it.  
We define
\[
\mathrm{SDS} = \frac{1}{|V|} \sum_{v \in V} \bigl(|f(v)| - 1\bigr).
\]
By definition, \(\mathrm{SDS} = 0\) if every concept is consistently mapped to a single symbol (no drift).  
When \(\mathrm{SDS} > 0\), symbol drift occurs, and higher values indicate stronger inconsistency in symbol usage.

\subsection{Main Result}

\begin{figure}[t]
    \centering
    \includegraphics[width=0.9\linewidth]{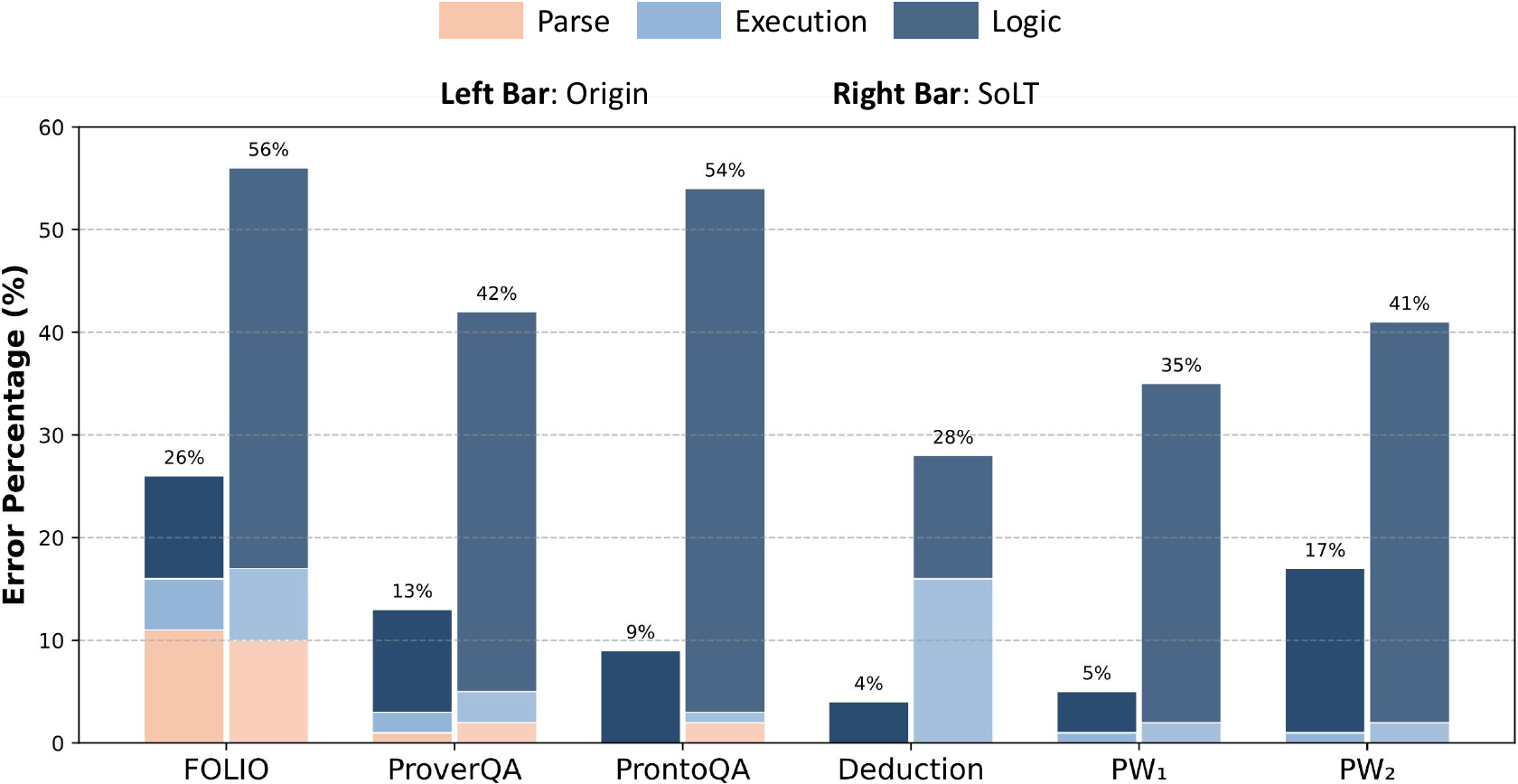} 
    \caption{Error statistics for GPT-4 across six tasks. }
    \label{fig:err-static}
\end{figure}

\begin{table*}[]
\centering
\caption{Accuracy(ACC, \%) and Symbol Dispersion Score (SDS) of four LLMs applying MenTaL via in-context learning across the original and SoLT diversified versions of six tasks. w/o stands for accuracy with direct prompting. w stands for accuracy with MenTaL. imp stands for improvement through MenTaL.}
\label{tab:main-results}
\resizebox{.98\textwidth}{!}{
\begin{tabular}{lccccccccccccccccccccc}
\toprule
\multicolumn{1}{c|}{\multirow{3}{*}{\textbf{Tasks}}} & \multicolumn{6}{c|}{GPT-3.5-Turbo}                                                                                                                                & \multicolumn{6}{c|}{GPT-4}                                                                                                                                       & \multicolumn{6}{c|}{DeepSeek-V3}                                                                                                                                 & \multicolumn{3}{c}{Llama}                                                 \\ \cmidrule(lr){2-22} 
\multicolumn{1}{c|}{}                                & \multicolumn{3}{c}{ACC $\uparrow$}                                                   & \multicolumn{3}{c|}{SDS $\downarrow$}                                                             & \multicolumn{3}{c}{ACC $\uparrow$}                                                   & \multicolumn{3}{c|}{SDS $\downarrow$}                                                             & \multicolumn{3}{c}{ACC $\uparrow$}                                                   & \multicolumn{3}{c|}{SDS $\downarrow$}                                                             & \multicolumn{3}{c}{ACC $\uparrow$}                                                   \\ \cmidrule(lr){2-22} 
\multicolumn{1}{c|}{}                                & \multicolumn{1}{c}{w/o} & \multicolumn{1}{c}{w} & \multicolumn{1}{c}{imp} & \multicolumn{1}{c}{w/o} & \multicolumn{1}{c}{w} & \multicolumn{1}{c|}{imp}           & \multicolumn{1}{c}{w/o} & \multicolumn{1}{c}{w} & \multicolumn{1}{c}{imp} & \multicolumn{1}{c}{w/o} & \multicolumn{1}{c}{w} & \multicolumn{1}{c|}{imp}           & \multicolumn{1}{c}{w/o} & \multicolumn{1}{c}{w} & \multicolumn{1}{c}{imp} & \multicolumn{1}{c}{w/o} & \multicolumn{1}{c}{w} & \multicolumn{1}{c|}{imp}           & \multicolumn{1}{c}{w/o} & \multicolumn{1}{c}{w} & \multicolumn{1}{c}{imp} \\ \midrule
\multicolumn{22}{c}{\textit{Origin}}                                                                                                                                                                                                                                                                                                                                                                                                                                                                                                                                                                                                               \\ \midrule
\multicolumn{1}{l|}{FOLIO}                           & 48.83                 & 59.57                   & \textbf{10.74}          & 0.10                  & 0.02                    & \multicolumn{1}{l|}{\textbf{0.08}} & 73.83                 & 78.17                   & \textbf{4.34}           & 0.02                  & 0.01                    & \multicolumn{1}{l|}{\textbf{0.01}} & 72.45                 & 76.41                   & \textbf{3.96}           & 0.05                  & 0.00                    & \multicolumn{1}{l|}{\textbf{0.05}} & 42.86                 & 23.99                   & -18.87                  \\
\multicolumn{1}{l|}{ProverQA}                        & 70.90                 & 75.33                   & 4.43                    & 0.01                  & 0.01                    & \multicolumn{1}{l|}{0.00}          & 86.84                 & 88.10                   & \textbf{1.26}           & 0.01                  & 0.00                    & \multicolumn{1}{l|}{\textbf{0.01}} & 85.83                 & 89.78                   & \textbf{3.95}           & 0.05                  & 0.04                    & \multicolumn{1}{l|}{\textbf{0.01}} & 61.76                 & 44.90                   & -16.86                  \\
\multicolumn{1}{l|}{ProntoQA}                        & 77.21                 & 91.14                   & \textbf{13.93}          & 0.06                  & 0.03                    & \multicolumn{1}{l|}{\textbf{0.03}} & 90.50                 & 98.02                   & \textbf{7.52}           & 0.06                  & 0.01                    & \multicolumn{1}{l|}{\textbf{0.05}} & 87.83                 & 98.01                   & \textbf{10.18}          & 0.07                  & 0.05                    & \multicolumn{1}{l|}{\textbf{0.02}} & 57.14                 & 70.27                   & \textbf{13.13}          \\
\multicolumn{1}{l|}{Deduction}                       & 69.06                 & 70.33                   & \textbf{1.27}           & 0.00                  & 0.00                    & \multicolumn{1}{l|}{0.00}          & 95.99                 & 96.45                   & \textbf{0.46}           & 0.00                  & 0.00                    & \multicolumn{1}{l|}{\textbf{0.00}} & 89.63                 & 96.62                   & \textbf{6.99}           & 0.00                  & 0.00                    & \multicolumn{1}{l|}{0.00}          & 54.52                 & 49.33                   & -5.19                   \\
\multicolumn{1}{l|}{PW1}                             & 94.17                 & 94.84                   & \textbf{0.67}           & 0.01                  & 0.01                    & \multicolumn{1}{l|}{0.00}          & 95.48                 & 97.86                   & \textbf{2.38}           & 0.01                  & 0.00                    & \multicolumn{1}{l|}{\textbf{0.01}} & 99.17                 & 99.03                   & -0.14                   & 0.00                  & 0.00                    & \multicolumn{1}{l|}{0.00}          & 58.62                 & 75.65                   & \textbf{17.03}          \\
\multicolumn{1}{l|}{PW2}                             & 71.62                 & 76.47                   & \textbf{4.85}           & 0.01                  & 0.00                    & \multicolumn{1}{l|}{\textbf{0.01}} & 83.38                 & 83.72                   & \textbf{0.34}           & 0.01                  & 0.00                    & \multicolumn{1}{l|}{\textbf{0.01}} & 82.81                 & 83.76                   & \textbf{0.95}           & 0.01                  & 0.00                    & \multicolumn{1}{l|}{\textbf{0.01}} & 65.22                 & 72.86                   & \textbf{7.64}           \\ \midrule
\multicolumn{22}{c}{\textit{SoLT}}                                                                                                                                                                                                                                                                                                                                                                                                                                                                                                                                                                                                                \\ \midrule
\multicolumn{1}{l|}{FOLIO}                           & 12.17                 & 27.66                   & \textbf{15.49}          & 0.68                  & 0.21                    & \multicolumn{1}{l|}{\textbf{0.47}} & 44.17                 & 68.33                   & \textbf{24.16}          & 0.34                  & 0.05                    & \multicolumn{1}{l|}{\textbf{0.29}} & 22.12                 & 61.25                   & \textbf{39.13}          & 0.50                  & 0.03                    & \multicolumn{1}{l|}{\textbf{0.47}} & 10.47                 & 9.52                    & -0.95                   \\
\multicolumn{1}{l|}{ProverQA}                        & 47.01                 & 66.75                   & \textbf{19.74}          & 0.41                  & 0.08                    & \multicolumn{1}{l|}{\textbf{0.32}} & 58.19                 & 76.72                   & \textbf{18.53}          & 0.40                  & 0.17                    & \multicolumn{1}{l|}{\textbf{0.23}} & 50.86                 & 79.71                   & \textbf{28.85}          & 0.49                  & 0.11                    & \multicolumn{1}{l|}{\textbf{0.38}} & 44.54                 & 37.12                   & -7.42                   \\
\multicolumn{1}{l|}{ProntoQA}                        & 44.50                 & 68.60                   & \textbf{24.10}          & 0.64                  & 0.10                    & \multicolumn{1}{l|}{\textbf{0.54}} & 46.39                 & 87.13                   & \textbf{40.74}          & 0.46                  & 0.06                    & \multicolumn{1}{l|}{\textbf{0.40}} & 57.19                 & 94.13                   & \textbf{36.94}          & 0.49                  & 0.06                    & \multicolumn{1}{l|}{\textbf{0.43}} & 37.21                 & 59.57                   & \textbf{22.36}          \\
\multicolumn{1}{l|}{Deduction}                       & 25.33                 & 39.27                   & \textbf{13.94}          & 0.69                  & 0.09                    & \multicolumn{1}{l|}{\textbf{0.60}} & 71.98                 & 82.79                   & \textbf{10.81}          & 0.59                  & 0.04                    & \multicolumn{1}{l|}{\textbf{0.55}} & 59.06                 & 80.46                   & \textbf{21.40}          & 0.79                  & 0.05                    & \multicolumn{1}{l|}{\textbf{0.74}} & 14.29                 & 12.59                   & -1.70                   \\
\multicolumn{1}{l|}{PW1}                             & 48.03                 & 73.81                   & \textbf{25.78}          & 0.35                  & 0.10                    & \multicolumn{1}{l|}{\textbf{0.25}} & 65.38                 & 91.76                   & \textbf{26.38}          & 0.30                  & 0.04                    & \multicolumn{1}{l|}{\textbf{0.26}} & 53.01                 & 87.20                   & \textbf{34.19}          & 0.35                  & 0.02                    & \multicolumn{1}{l|}{\textbf{0.32}} & 42.33                 & 47.14                   & \textbf{4.81}           \\
\multicolumn{1}{l|}{PW2}                             & 42.12                 & 60.49                   & \textbf{18.37}          & 0.35                  & 0.15                    & \multicolumn{1}{l|}{\textbf{0.19}} & 58.80                 & 78.57                   & \textbf{19.77}          & 0.38                  & 0.03                    & \multicolumn{1}{l|}{\textbf{0.35}} & 50.67                 & 77.89                   & \textbf{27.22}          & 0.44                  & 0.02                    & \multicolumn{1}{l|}{\textbf{0.42}} & 34.83                 & 50.41                   & \textbf{15.58}          \\ \bottomrule
\end{tabular}}
\end{table*}

Using the above experimental setup, we evaluated four LLMs across six datasets with different solver configurations.  
Figure~\ref{fig:multi-model-accuracy} shows their accuracy in solving logical problems under various tasks and prompting strategies.

\paragraph{Inconsistent Symbol Mapping under Linguistic Diversification.}
Under direct prompting, accuracy drops substantially after SoLT diversification, from \texttt{0.51} to \texttt{0.24}.
This drop coincides with a sharp increase in the Symbol Dispersion Score (SDS): SDS remains near \texttt{0} on original tasks but rises significantly on diversified inputs.
These results indicate that LLMs often assign different logical symbols to semantically equivalent expressions, leading to symbol drift and degraded reasoning performance.


\paragraph{Prompt-tuning Fails to Enable Consistent Symbol Mapping.}
In the prompt-tuning setting, we explicitly instructed LLMs to align identical concepts with consistent logical symbols and provided in-context examples showing diverse expressions of the same concept.  
However, when comparing prompt-tuning with direct prompting on SoLT-diversified tasks, we found no significant improvement in either SDS or overall accuracy.  
This suggests that minor prompt adjustments, even with additional examples, are still insufficient to help LLMs maintain consistent symbol mapping across varied expressions, and thus cannot reliably improve their performance on diversified logical reasoning tasks.

\subsection{Response Pattern Analysis}

\label{sec:response}

\begin{figure}[t]
    \centering
    \includegraphics[width=0.45\textwidth]{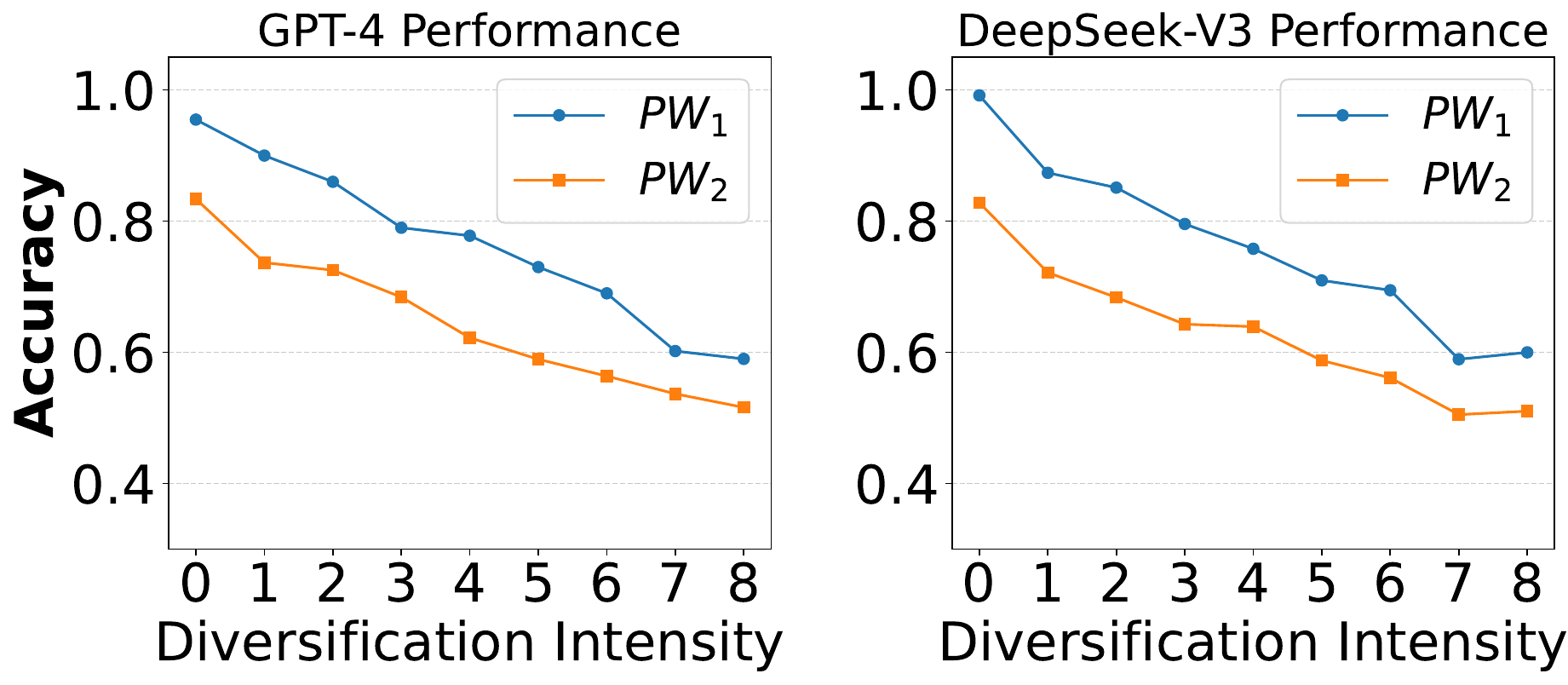} 
    \caption{The performance of GPT-4 and DeepSeek-V3 on PW$_1$ and PW$_2$ varies with Diversification Intensity.}
    \label{fig:zhexian}
\end{figure}

We further analyze GPT-4 outputs by categorizing translation errors into three types (Figure~\ref{fig:err-static}): parsing errors, execution errors, and logic errors.
Overall, SoLT diversification does not substantially increase parsing or execution errors across most tasks, indicating that translation executability remains largely stable.
In contrast, logic errors rise sharply, suggesting that linguistic diversification primarily challenges the model’s ability to maintain logical consistency rather than syntactic correctness.
An exception occurs in the Deduction dataset, where execution errors increase due to the \texttt{python-constraint} solver’s requirement that all variables be defined before use, which exposes cases where the model introduces undefined symbols by assigning different identifiers to diversified expressions of the same concept.

\subsection{Performance on Diversification Intensity}


We further adjusted the SoLT strategy to systematically control the frequency of linguistic diversification and conducted experiments on the ProofWriter dataset, integrating two symbolic solvers with GPT-4 and DeepSeek-V3.
Specifically, diversification intensity is controlled by progressively increasing the number of sentences in each reasoning instance that undergo SoLT-based replacement, while others remain unchanged.
The results, shown in Figure~\ref{fig:zhexian}, demonstrate that as diversification intensity increases, the translation accuracy of LLMs decreases smoothly and continuously, revealing a clear cumulative effect.  
Specifically, with stronger diversification, LLMs must process a growing number of semantic variants, leading to a gradual buildup of translation errors.  
During this process, words and expressions essential for correct reasoning are progressively modified, making it increasingly difficult for LLMs to maintain consistent interpretation and symbol alignment, ultimately resulting in steadily declining performance.  
These findings indicate that LLMs have only limited ability to maintain consistent symbol mapping in formal logic translation, and that this limitation becomes increasingly pronounced under higher diversification.  
Moreover, diversification intensity can serve as a continuous metric for evaluating the stability of symbol consistency in LLMs.
\section{Enhancing LLM Translators via MenTaL}


\subsection{Guiding through In-Context Learning}

\begin{table}[t]
  \centering
  \caption{Accuracy (ACC, \%) and Symbol Dispersion Score (SDS) of LLaMA-3-8B-Instruct on original and SoLT tasks.  
Base: unfine-tuned model; SFT: fine-tuned without MenTaL; MenTaL: fine-tuned with MenTaL.  
Best results are in bold.}
  \label{tab:sft-results}
  \resizebox{0.95\linewidth}{!}{
\begin{tabular}{lcccccc}
\toprule
\multicolumn{1}{l|}{\multirow{2}{*}{Tasks}} & \multicolumn{3}{c|}{ACC $\uparrow$}                         & \multicolumn{3}{c}{SDS $\downarrow$}     \\ \cmidrule{2-7} 
\multicolumn{1}{c|}{}                       & Base & SFT  & \multicolumn{1}{c|}{MenTaL}        & Base & SFT  & MenTaL        \\ \midrule
\multicolumn{7}{c}{\textit{Origin}}                                                                                                   \\ \midrule
\multicolumn{1}{l|}{FOLIO}                  & 42.86 & 55.09 & \multicolumn{1}{c|}{\textbf{58.26}} & 0.13 & 0.10 & \textbf{0.06} \\
\multicolumn{1}{l|}{ProverQA}               & 61.76 & 75.55 & \multicolumn{1}{c|}{\textbf{79.16}} & 0.03 & 0.04 & \textbf{0.01} \\
\multicolumn{1}{l|}{ProntoQA}               & 57.14 & 68.75 & \multicolumn{1}{c|}{\textbf{82.42}} & 0.08 & 0.04 & \textbf{0.03} \\
\multicolumn{1}{l|}{Deduction}              & 54.52 & 62.37 & \multicolumn{1}{c|}{\textbf{67.54}} & 0.07 & 0.05 & \textbf{0.04} \\
\multicolumn{1}{l|}{PW$_1$}                    & 58.62 & 82.96 & \multicolumn{1}{c|}{\textbf{84.10}} & 0.11 & 0.05 & \textbf{0.04} \\
\multicolumn{1}{l|}{PW$_2$}                    & 65.22 & 73.17 & \multicolumn{1}{c|}{\textbf{75.20}} & 0.20 & 0.02 & \textbf{0.01} \\ \midrule
\multicolumn{7}{c}{\textit{SoLT}}                                                                                                    \\ \midrule
\multicolumn{1}{l|}{FOLIO}                  & 10.47 & 12.53 & \multicolumn{1}{c|}{\textbf{40.43}} & 0.78 & 0.64 & \textbf{0.13} \\
\multicolumn{1}{l|}{ProverQA}               & 44.54 & 51.39 & \multicolumn{1}{c|}{\textbf{67.21}} & 0.65 & 0.44 & \textbf{0.08} \\
\multicolumn{1}{l|}{ProntoQA}               & 37.21 & 44.73 & \multicolumn{1}{c|}{\textbf{75.67}} & 0.69 & 0.67 & \textbf{0.04} \\
\multicolumn{1}{l|}{Deduction}              & 14.29 & 16.44 & \multicolumn{1}{c|}{\textbf{38.78}} & 0.73 & 0.63 & \textbf{0.09} \\
\multicolumn{1}{l|}{PW$_1$}                    & 42.33 & 55.10 & \multicolumn{1}{c|}{\textbf{72.66}} & 0.39 & 0.35 & \textbf{0.10} \\
\multicolumn{1}{l|}{PW$_2$}                    & 34.83 & 47.83 & \multicolumn{1}{c|}{\textbf{69.71}} & 0.73 & 0.37 & \textbf{0.06} \\ \bottomrule
\end{tabular}}
\end{table}

We apply \textbf{MenTaL} via \textbf{in-context learning} to guide LLMs toward stable logical translation, by providing demonstrations constructed with mental representation tables.
This enables models to learn consistent predicate assignment for semantically equivalent expressions.
We evaluate four LLMs on both the original and SoLT-diversified versions of six tasks, with results shown in Table~\ref{tab:main-results}.

For high-performance models (GPT-3.5-Turbo, GPT-4, and DeepSeek-V3), applying MenTaL to the original datasets yields only marginal accuracy gains.
Consistently, SDS remains near \texttt{0} even without MenTaL, indicating minimal symbol drift due to limited linguistic diversity in the original data.
In contrast, on SoLT-diversified datasets, MenTaL leads to substantial improvements, with accuracy gains of up to \texttt{40.74}\% and a pronounced reduction in SDS, approaching zero.
These results show that MenTaL effectively enforces consistent symbol mapping under linguistic variation when applied via in-context learning.

For the smaller open-source \texttt{LLaMA-3-8B-Instruct} model, MenTaL improves accuracy on some datasets but decreases it on others. 
Further analysis shows that smaller models often fail to follow in-context demonstrations, leading to formatting issues and frequent parse or execution errors. 
This suggests that their limited capacity constrains reliable acquisition of stable symbol alignment through in-context learning, which warrants further investigation.

\subsection{Refining through Supervised Fine-Tuning}

To improve symbol consistency in small open-source LLMs, we adopt a supervised fine-tuning (SFT) approach~\cite{wu2024instruction,caciularu2024tact,zheng2024llamafactory}.
We construct a training set from successful reasoning traces generated by high-performance LLMs using MenTaL under in-context learning, and fine-tune the Llama-3-8B-Instruct model on this data.
For comparison, we train a baseline SFT model using outputs from the same LLMs without MenTaL, together with the original Llama-3-8B-Instruct model.
MenTaL-based SFT is implemented with LoRA; training details and costs are provided in Appendix~\ref{app:exp_set}.

On standard datasets, the non-MenTaL SFT model improves accuracy over the original model.
However, on SoLT-diversified datasets, its SDS increases markedly, indicating persistent symbol drift.
In contrast, the MenTaL-based SFT model not only attains the highest accuracy on the original datasets but also substantially outperforms both baselines on diversified ones, achieving near-zero SDS and markedly higher accuracy.  
These results demonstrate that incorporating MenTaL into fine-tuning enables small-scale open-source LLMs to maintain consistent symbol mapping and sustain stable performance under linguistically diversified conditions.

Notably, our MenTaL-based SFT was conducted only once yet covered three solver-specific formats and five dataset tasks, highlighting its strong generalization in embedding consistent symbol mapping across solvers and tasks.

\subsection{Error Attribution Analysis of MenTaL}

\begin{figure}[t]
    \centering
    \includegraphics[width=0.95\linewidth]{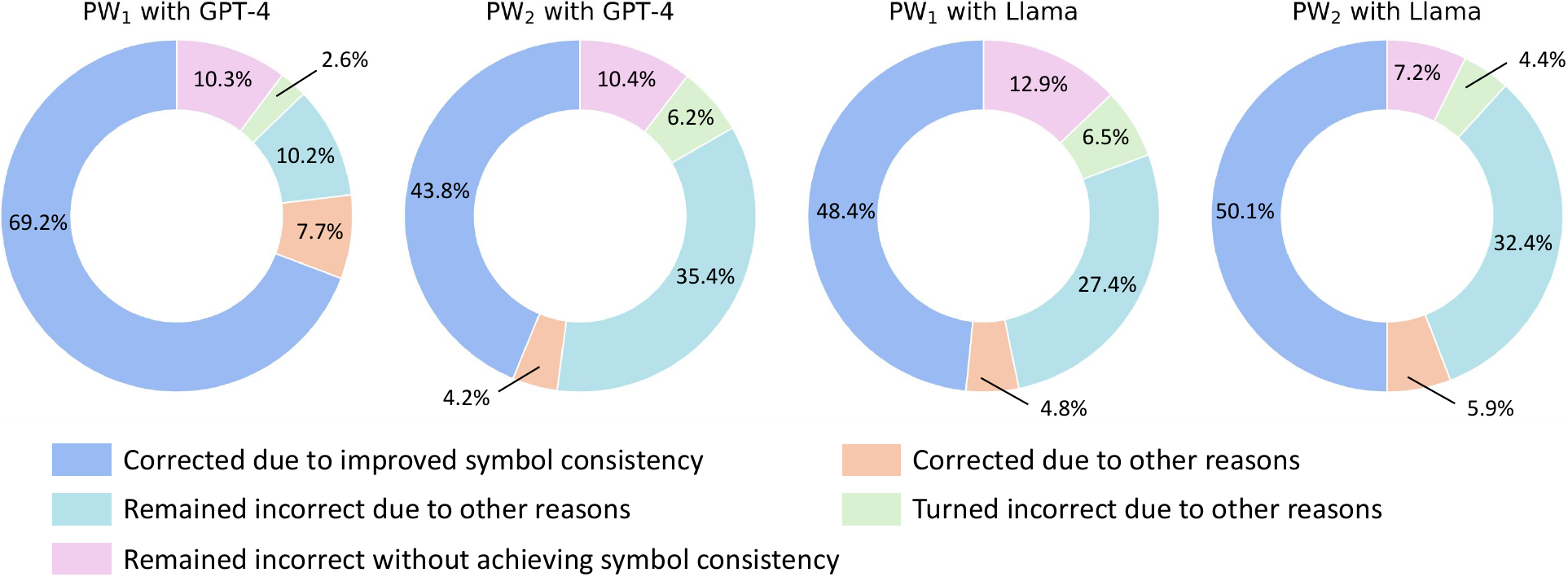} 
    \caption{Error transformation distribution before and after applying MenTaL on PW$_1$ and PW$_2$, showing proportions of corrections and remaining errors across different causes.}
    \label{fig:huan}
\end{figure}

To verify that MenTaL’s performance gains stem from improved symbol consistency, we conduct a manual error analysis in two representative settings: GPT-4 with prompt-based MenTaL and LLaMA with SFT-based MenTaL, evaluated on the SoLT-diversified PW$_1$ and PW$_2$ tasks.
Results are shown in Figure~\ref{fig:huan}.
We analyze errors that persist both before and after applying MenTaL, and categorize them into five types: corrections due to improved symbol consistency, corrections due to other factors, errors remaining for other reasons, errors newly introduced for other reasons, and errors remaining without achieving symbol consistency.
Here, “other reasons” denote factors unrelated to symbol mapping, such as limitations in translating logical relations.

Overall, the results show that about half of the errors were corrected through improved symbol consistency, while most of the remaining errors were caused by “other reasons,” namely the model’s limited ability in logical relation translation rather than inconsistency in mapping.  
This confirms that MenTaL’s performance gains mainly stem from its targeted enhancement of consistent symbol mapping, thereby improving the stability of both closed-source and open-source models under linguistic variation.
\section{Conclusion}
\label{sec:conclusion}

Linguistic diversity is a common feature of natural language, yet existing logical reasoning benchmarks rarely require models to handle such variation.  
We propose SoLT, which introduces multi-level linguistic diversification while preserving the original problem logic, allowing evaluation of models’ ability to process diversified text in logical reasoning.  
Using SoLT, we find that LLMs, when serving as logical translators, experience a significant accuracy drop under linguistic diversification.  
Further analysis shows that the main cause is their inability to maintain consistent symbol mapping, revealing a key deficiency overlooked by current benchmarks.  
The goal of this study is not only to identify these limitations but also to improve model capability.


\section{Acknowledgments}
This research was supported by grants from the National Key Research and Development Program of China (Grant No. 2024YFC3308 200), the National Natural Science Foundation of China (Nos. 625256 06 and 62502486), the Key Technologies R \& D Program of Anhui Province (No. 202423k09020039), and the Fundamental Research Funds for the Central Universities.

\bibliographystyle{ACM-Reference-Format}
\bibliography{sample-base}

\appendix


\section{Evaluation and Logical Invariance Validation}
\label{app:human}

\subsection{Annotation and Evaluation Details}

To verify that linguistic diversification preserves the original semantics and reasoning structure, we conducted a human and model-based evaluation following the logical invariance criteria in Table~2.

\paragraph{Annotators.}
Three PhD-level annotators with expertise in NLP and formal reasoning independently evaluated the data. None participated in the construction of the SoLT benchmark, ensuring an unbiased assessment.

\paragraph{Data sampling.}
For each dataset, we randomly sampled 100 instances. Annotators were presented with paired original and SoLT-diversified versions of each instance to enable direct comparison focused on semantic and logical preservation.

\paragraph{Aggregation and reliability.}
Final human scores were obtained by averaging across annotators. Inter-annotator agreement is high (Cohen’s $\kappa = 0.83$), indicating reliable judgments.

\paragraph{Model-based evaluation.}
We additionally report LLM-based evaluations using the same criteria. The logical invariance scores in Table~1 are computed by averaging human and model judgments, providing a complementary evaluation perspective.

\subsection{Human Evaluation of SoLT Diversification}

\begin{table}[h]
    \centering
    \caption{Contingency Table of Human Performance on Original and SoLT Datasets}
    \label{tab:contingency}
    \resizebox{0.65\linewidth}{!}{
    \begin{tabular}{c|cc|c}
        \toprule
        \diagbox{SoLT}{Origin} & Right & Wrong & Total \\
        \hline
        Right & 184 & 4 & 188 \\
        Wrong & 8 & 4 & 12 \\
        \hline
        Total & 196 & 4 & 200 \\
        \bottomrule
    \end{tabular}}
\end{table}

We also conducted a more direct experiment, recruiting human participants to answer randomly selected questions from five original datasets and their SoLT-diversified counterparts. To ensure unbiased comparison, we adopted an A/B testing setup: each participant was shown either the original or the SoLT version of a question, but never both. The results are shown in Table \ref{tab:contingency}. Human performance on both versions was excellent, approaching perfect scores, demonstrating that \textbf{diversification in SoLT preserve human interpretability and do not hinder problem-solving.}

\section{Algorithm for MenTaL}
\begin{algorithm}[]
\caption{MenTaL: Mental Representation Table-Guided Logical Translation}
\label{alg:mental}
\KwIn{Logical problem $q = \{e_1, e_2, \dots, e_n\}$}
\KwOut{Logical form $\tilde{q}$ and Mental Representation Table $T$}

\BlankLine
Initialize $\tilde{q} \leftarrow \emptyset$, $T \leftarrow \emptyset$\;
\ForEach{expression $e^{(i)}$ in $q$}{
    \ForEach{entry $(E_j, \sigma_j)$ in $T$}{
        \If{\textsc{LLMEquiv}$(e^{(i)}, E_j)$}{
            Assign symbol $\sigma(e^{(i)}) \leftarrow \sigma_j$\;
            Append $\sigma(e^{(i)})$ to $\tilde{q}$\;
            \KwContinue\;
        }
        \If{\textsc{LLMConflict}$(e^{(i)}, E_j)$}{
            Select more atomic concept $e^{*}$ between $e^{(i)}$ and some $e^{(k)} \in E_j$\;
            Rewrite $\sigma(e^{(i)}) \leftarrow \sigma(e^{*}) \wedge \sigma(\text{modifier})$\;
            Retroactively update prior occurrences in $\tilde{q}$\;
            Update $T \leftarrow T \cup \{(E_j \cup \{e^{(i)}\}, \sigma(e^{(i)}))\}$\;
            \KwContinue\;
        }
    }
    Create new symbol $\sigma_{\text{new}}$\;
    Add new entry $(\{e^{(i)}\}, \sigma_{\text{new}})$ to $T$\;
    Append $\sigma_{\text{new}}$ to $\tilde{q}$\;
}
\Return $(\tilde{q}, T)$
\end{algorithm}
Algorithm~\ref{alg:mental} introduces several symbols and routines not defined in the main text.  
Given a natural-language problem $q=\{e_1,e_2,\dots,e_n\}$, each $e_i$ denotes an \emph{expression}-level unit 
(word, phrase, or clause), distinct from the sentence-level notation $s_i$ in Section~4.  
$\tilde{q}$ is the logical form incrementally generated, and $T$ denotes the \emph{Mental Representation Table}, 
mapping semantically equivalent expressions to a unified symbol:  
$T:\{e^{(1)},e^{(2)},\dots,e^{(m)}\}\mapsto\sigma$.  
$\sigma(e_i)$ is the symbol assigned to $e_i$. $\sigma_{\text{new}}$ denotes a newly created symbol.

Two LLM-based functions are defined:  
$\textsc{LLMEquiv}(e,E_j)$ determines whether $e$ and any element of $E_j$ are semantically equivalent,  
while $\textsc{LLMConflict}(e,E_j)$ detects if $e$ modifies or specializes an existing concept.  
When a conflict occurs, the algorithm keeps the more atomic concept~$e^*$ and rewrites  
$\sigma(e^{(i)})\!\leftarrow\!\sigma(e^{*})\wedge\sigma(\text{modifier})$, preserving compositional consistency.  
When $T$ is updated, earlier parts of $\tilde{q}$ are retroactively revised to maintain symbol coherence.

Together, the three operations—\emph{Reusing}, \emph{Extending}, and \emph{Conflict Resolution}—maintain concept-level symbol alignment, 
preventing symbol drift and enforcing consistent mapping across semantically equivalent expressions.

\section{Experiment Details}
\label{app:exp_set}
\subsection{Details of Datasets}
\label{datasets-set}

\textbf{FOLIO.}  
An expert-curated open-domain dataset for natural language reasoning with first-order logic, featuring diverse structures and high logical complexity.  
We use the full validation set of 130 examples.

\textbf{ProofWriter.}  
An automatically generated dataset where each example consists of a passage and a conclusion, and the task is to determine whether the conclusion logically follows.  
We randomly sample 500 examples with reasoning depths from 1 to 5.

\textbf{Deduction.}  
A subset of BigBench focusing on constraint-based reasoning.  
Each problem includes a context, a question, and multiple options, with the goal of identifying all options consistent with the stated constraints.  
We sample 200 examples from each of its 3-, 5-, and 7-option variants.

\textbf{ProntoQA.}  
A rule-based dataset with explicit reasoning chains.  
Each example includes a context, a query, and a chain of thought.  
Attributes are single-word tokens, often artificial rather than standard English.  
We randomly select 200 examples.

\textbf{ProverQA.}  
Derived from ProofWiki, this dataset contains formalized proofs from mathematical texts.  
Each problem includes a context, a question, and an extracted proof chain.  
We use 200 examples from the \textit{medium} difficulty level, comparable to FOLIO.

Table~\ref{tab:testset_sizes} summarizes the sample counts for each dataset.

\begin{table}[h]
\centering
\caption{Number of examples used in final test sets.}
\label{tab:testset_sizes}
\resizebox{0.95\linewidth}{!}{
\begin{tabular}{l c c c c c}
\toprule
\textbf{Dataset} & FOLIO & ProofWriter & Deduction & ProntoQA & ProverQA \\
\midrule
\textbf{\# Examples} & 130 & 500 & 200 & 200 & 200 \\
\bottomrule
\end{tabular}}
\end{table}

\subsection{Details of Solvers}

\textbf{Python-Constraint.}  
A declarative solver for constraint satisfaction problems (CSP), supporting the definition of variables, domains, and logical constraints.

\textbf{PyKE (Python Knowledge Engine).}  
A rule-based reasoning framework combining logic programming with object-oriented design, suitable for decision and inference systems.

\textbf{Prover9.}  
An automated theorem prover for first-order and equational logic.

Each dataset is paired with a specific solver: ProofWriter supports both \textbf{Prover9} and \textbf{PyKE}, while all others use a single solver.  
The mapping is summarized in Table~\ref{tab:solver_model}.

\begin{table}[h]
\centering
\caption{Mapping Between Datasets and Solvers}
\label{tab:solver_model}
\resizebox{0.95\linewidth}{!}{
\begin{tabular}{l c c c c c}
\toprule
\textbf{Dataset} & Deduction & ProntoQA & FOLIO & ProofWriter & ProverQA \\
\midrule
\textbf{Solver(s)} & Python-Constraint & PyKE & Prover9 & PyKE, Prover9 & Prover9 \\
\bottomrule
\end{tabular}}
\end{table}

\subsection{Symbol Dispersion Score (SDS) Computation}

SDS exploits a key property of SoLT: for each instance, linguistic variants referring to the same semantic concept are preserved and identifiable.
For each diversified sample, expressions corresponding to the same concept are grouped according to the SoLT construction.
After logical translation, the symbol assigned to each expression is identified, with GPT-4 used to align expressions with their translated symbols.
For each concept $v$, we count the number of distinct symbols across its variants, denoted as $|f(v)|$.
SDS is then computed by averaging $|f(v)| - 1$ over all concepts in the dataset.
All statistics are manually verified to correct potential alignment errors.


\subsection{Evaluation Setting}
\label{app-D-2}
\paragraph{Model and Environment Settings.}
We evaluate both close-sourced (GPT-4, GPT-3.5-Turbo, DeepSeek-V3, DeepSeek-R1) and open-sourced (LLaMA-3-8B-Instruct) LLMs. Close-sourced models are accessed via APIs with the temperature set to 0.2, while open-sourced models are locally deployed using their HuggingFace versions with default settings. All experiments are conducted on a Linux server equipped with Intel Xeon Platinum 8269CY CPUs and one NVIDIA A100 GPU (40GB). The software environment includes Python 3.10, PyTorch 2.2.0, and Transformers 4.39.3.

\paragraph{Prompt Setting.}
We adopt prompt templates to construct evaluation queries. The Direct prompting strategy follows the Logic-LM setting. Model outputs are transformed into JSON format for analysis, with regular expressions applied to improve parsing robustness. Further details are provided in the codebase.

\subsection{SFT Setting}
To study the impact of \textbf{MenTaL}-guided translation on model learning, we conduct two supervised fine-tuning (SFT) runs under identical settings using \textbf{LLaMA-Factory} with LoRA-based tuning.

\paragraph{Data Setup.}
For the \textbf{baseline SFT}, training data are logical translations of the original datasets generated by \textit{GPT-4} and \textit{DeepSeek-V3}.  
For the \textbf{MenTaL SFT}, we sample 600 instances from six reasoning tasks and generate MenTaL-guided logical translations via in-context learning with \textit{GPT-4} and \textit{DeepSeek-V3}.  
Both settings use the same training size (600 examples). Models are fine-tuned once and evaluated on the test sets of all six tasks (Table~\ref{tab:sft-results}).

\paragraph{Fine-tuning Method.}
We employ \textbf{LoRA} for parameter-efficient tuning, updating only adapter parameters while keeping the base model frozen.

\paragraph{Training Configuration.}  
Both SFT settings share the same hyperparameters:
\begin{itemize}
    \item \textbf{Batch size:} 1 (given the small dataset size of 600 samples);  
    \item \textbf{Learning rate:} 5e-5;  
    \item \textbf{Epochs:} 3;  
    \item \textbf{Optimizer:} AdamW with default parameters;  
    \item \textbf{Precision:} FP16 mixed precision for GPU efficiency.
\end{itemize}

\paragraph{Environment.}  
All fine-tuning experiments are conducted on a Linux server same as the one menthioned in~\ref{app-D-2}.

\paragraph{Comparison Protocol.}  
After fine-tuning, both models are evaluated under identical experimental conditions.  
This setup isolates the influence of \textbf{MenTaL-guided translation}, allowing us to assess how training on SoLT-diversified yet logically equivalent data improves semantic consistency and reasoning accuracy compared to the baseline SFT.

\section{Experiment Cost}
\label{app-E}

In our experiments, LLM API usage (with GPT-4 as an example) covers four components:
\textbf{E0} SoLT dataset generation,
\textbf{E1} accuracy evaluation on Original and SoLT datasets under two prompting strategies,
\textbf{E2} perturbation generation and evaluation on ProofWriter, and
\textbf{E3} MenTaL with in-context learning.

Token consumption and corresponding GPT-4 costs are reported in Tables~\ref{tab:token_usage} and~\ref{tab:token_cost}, respectively.

\paragraph{Additional Cost of MenTaL}
Compared to Direct translation, MenTaL increases token usage by \textbf{1.3}$\times$ for input tokens and \textbf{2.1}$\times$ for output tokens.

\begin{table}[h]
\centering
\caption{Token Usage Statistics for Different Experiment Stages (GPT-4)}
\label{tab:token_usage}

\begin{tabular}{lcc}
\toprule
\textbf{Index} & \textbf{\# Input Tokens} & \textbf{\# Output Tokens} \\
\midrule
E0 & 1,658,015 & 354,493 \\
E1 & 9,765,819 & 2,113,512 \\
E2 & 4,880,444 & 1,023,106 \\
E3 & 5,040,127 & 2,646,814 \\
\bottomrule
\end{tabular}
\end{table}

\begin{table}[h]
\centering
\caption{Estimated Cost (USD) for Token Usage in Different Experiment Stages (GPT-4)}
\label{tab:token_cost}
\resizebox{0.9\linewidth}{!}{
\begin{tabular}{lccc}
\toprule
\textbf{Index} & \textbf{Input Cost (\$)} & \textbf{Output Cost (\$)} & \textbf{Total Cost (\$)} \\
\midrule
E0 & 16.58 & 10.63 & 27.21 \\
E1 & 97.65 & 63.43 & 161.08 \\
E2 & 48.80 & 30.69 & 79.49 \\
E3 & 50.40 & 79.40 & 129.80 \\
Total & 213.43     & 184.15     & 397.58 \\
\bottomrule
\end{tabular}}
\end{table}

\section{New Baselines}
\label{appendix:a}

For comparison, we include two recent LLM-based logical translation frameworks. Since their results on the original benchmarks have already been reported in the respective papers, we conduct experiments only on the SoLT-diversified datasets.

\textbf{Divide-and-Translate}~\cite{ryu2024divide} adopts a compositional approach that decomposes problems into semantic subunits, translates them independently, and recombines them into a full logical form.  
While this improves structural fidelity, it does not unify semantically equivalent expressions across the problem.

\textbf{Logic-LM++}~\cite{kirtania2024logic} extends Logic-LM with multi-step refinement, iteratively revising and verifying an initial translation through structured prompts.  
Although this enhances local accuracy, it lacks a mechanism to enforce global symbol consistency.

Based on Table~\ref{tab:new}, clear trends emerge.  
Although \textbf{Divide-and-Translate} and \textbf{Logic-LM++} outperform direct translation in their original settings, these gains largely disappear under SoLT.  
\textbf{Logic-LM++} shows only marginal improvement over Direct, while \textbf{Divide-and-Translate} even degrades in accuracy.  
This suggests that self-checking in \textbf{Logic-LM++} cannot effectively correct symbol drift, and that the sentence-level decomposition in \textbf{Divide-and-Translate}, followed by only a single global check, is insufficient to maintain symbol consistency under linguistic diversity.

\begin{table}[t]
\centering
\caption{Performance of Recent Logical-Translation Baselines on SoLT-Diversified Tasks}
\label{tab:new}
\resizebox{\linewidth}{!}{
\begin{tabular}{lcccccccc}
\toprule
\multicolumn{1}{c|}{\multirow{2}{*}{\textbf{Tasks}}} & \multicolumn{2}{c|}{GPT-3.5-Turbo} & \multicolumn{2}{c|}{GPT-4}        & \multicolumn{2}{c|}{DeepSeek-V3}  & \multicolumn{2}{c}{DeepSeek-R1} \\ \cmidrule{2-9} 
\multicolumn{1}{c|}{}                                & ACC    & \multicolumn{1}{l|}{SDS}  & ACC   & \multicolumn{1}{l|}{SDS}  & ACC   & \multicolumn{1}{l|}{SDS}  & ACC             & SDS           \\ \midrule
\multicolumn{9}{c}{\textit{Divide and Translate in SoLT}}                                                                                                                                                            \\ \midrule
\multicolumn{1}{l|}{FOLIO}                           & 21.38  & \multicolumn{1}{l|}{0.72} & 37.66 & \multicolumn{1}{l|}{0.44} & 31.59 & \multicolumn{1}{l|}{0.46} & 32.87           & 0.43          \\
\multicolumn{1}{l|}{ProverQA}                        & 40.82  & \multicolumn{1}{l|}{0.52} & 50.73 & \multicolumn{1}{l|}{0.56} & 44.92 & \multicolumn{1}{l|}{0.58} & 73.19           & 0.30          \\
\multicolumn{1}{l|}{ProntoQA}                        & 37.64  & \multicolumn{1}{l|}{0.75} & 47.12 & \multicolumn{1}{l|}{0.45} & 39.47 & \multicolumn{1}{l|}{0.73} & 49.36           & 0.56          \\
\multicolumn{1}{l|}{Deduction}                       & 33.71  & \multicolumn{1}{l|}{0.57} & 56.88 & \multicolumn{1}{l|}{0.67} & 54.37 & \multicolumn{1}{l|}{0.69} & 59.41           & 0.59          \\
\multicolumn{1}{l|}{PW$_1$}                          & 42.55  & \multicolumn{1}{l|}{0.47} & 54.94 & \multicolumn{1}{l|}{0.52} & 50.83 & \multicolumn{1}{l|}{0.47} & 58.24           & 0.33          \\
\multicolumn{1}{l|}{PW$_2$}                          & 39.74  & \multicolumn{1}{l|}{0.40} & 49.85 & \multicolumn{1}{l|}{0.49} & 46.68 & \multicolumn{1}{l|}{0.50} & 54.37           & 0.38          \\ \midrule
\multicolumn{9}{c}{\textit{Logic-LM++ in SoLT}}                                                                                                                                                                      \\ \midrule
\multicolumn{1}{l|}{FOLIO}                           & 15.42  & \multicolumn{1}{l|}{0.60}  & 47.31 & \multicolumn{1}{l|}{0.32} & 24.28 & \multicolumn{1}{l|}{0.53} & 41.74           & 0.36          \\
\multicolumn{1}{l|}{ProverQA}                        & 48.63  & \multicolumn{1}{l|}{0.41} & 60.25 & \multicolumn{1}{l|}{0.39} & 51.19 & \multicolumn{1}{l|}{0.49} & 67.54           & 0.34          \\
\multicolumn{1}{l|}{ProntoQA}                        & 45.18  & \multicolumn{1}{l|}{0.64} & 47.77 & \multicolumn{1}{l|}{0.46} & 60.35 & \multicolumn{1}{l|}{0.47} & 54.69           & 0.50           \\
\multicolumn{1}{l|}{Deduction}                       & 27.39  & \multicolumn{1}{l|}{0.68} & 72.41 & \multicolumn{1}{l|}{0.56} & 63.82 & \multicolumn{1}{l|}{0.75} & 65.26           & 0.52          \\
\multicolumn{1}{l|}{PW$_1$}                          & 49.58  & \multicolumn{1}{l|}{0.35} & 66.22 & \multicolumn{1}{l|}{0.32} & 54.74 & \multicolumn{1}{l|}{0.34} & 60.87           & 0.33          \\
\multicolumn{1}{l|}{PW$_2$}                          & 43.44  & \multicolumn{1}{l|}{0.34} & 60.17 & \multicolumn{1}{l|}{0.37} & 53.28 & \multicolumn{1}{l|}{0.43} & 59.66           & 0.34          \\ \bottomrule
\end{tabular}}
\end{table}

\section{Semantic Proximity of SoLT Rewriting}
\label{sec:solt_semantic_proximity}

We evaluate whether synonym substitutions introduced by SoLT preserve sufficient semantic proximity.
For each rewritten concept pair, we compute cosine similarity using ConceptNet--Numberbatch embeddings.
Table~\ref{tab:conceptnet_similarity} reports the average similarity scores across datasets, all exceeding 0.90.

\begin{table}[]
\centering
\caption{Average cosine similarity between SoLT rewriting pairs measured by ConceptNet--Numberbatch embeddings.}
\label{tab:conceptnet_similarity}
\resizebox{0.95\linewidth}{!}{
\begin{tabular}{lccccc}
\toprule
Dataset & FOLIO & ProverQA & ProntoQA & Deduction & ProofWriter \\
\midrule
Similarity & 0.9160 & 0.9391 & 0.9645 & 0.9007 & 0.9312 \\
\bottomrule
\end{tabular}}
\end{table}

\section{SMT-Solver Consistency of MenTaL Translation}
\label{sec:smt_consistency}

We evaluate whether MenTaL improves logical translation consistency with respect to SMT solvers.
Using GPT-4 as the translator, we compare Direct translation and MenTaL-based translation by measuring consistency with gold-standard logical forms.
As shown in Table~\ref{tab:smt_consistency}, MenTaL yields substantially higher solver consistency across all datasets.

\begin{table}[]
\centering
\caption{SMT-solver consistency of logical translations produced by Direct and MenTaL-based methods.}
\label{tab:smt_consistency}
\resizebox{0.95\linewidth}{!}{
\begin{tabular}{lcccccc}
\toprule
Dataset & FOLIO & ProverQA & ProntoQA & Deduction & PW$_1$ & PW$_2$ \\
\midrule
MenTaL & 0.9055 & 0.8655 & 0.9172 & 0.8625 & 0.9558 & 0.9423 \\
Direct & 0.5823 & 0.6565 & 0.4987 & 0.7499 & 0.6810 & 0.7052 \\
\bottomrule
\end{tabular}}
\end{table}

\end{document}